\documentclass[10pt,twocolumn,letterpaper]{article}

\usepackage{iccv}              
\usepackage{multirow}
\usepackage{graphicx}
\usepackage{caption}
\usepackage{subcaption}

\definecolor{iccvblue}{rgb}{0.21,0.49,0.74}
\usepackage[pagebackref,breaklinks,colorlinks,allcolors=iccvblue]{hyperref}

\def\paperID{*****} 
\def\confName{ICCV}
\def\confYear{2025}

\title{Agentic Surgical AI: Surgeon Style Fingerprinting and Privacy Risk Quantification via Discrete Diffusion in a Vision-Language-Action Framework}

\author{
Huixin Zhan$^{\spadesuit}$, 
Jason H. Moore$^{\diamondsuit}$\thanks{Corresponding author} \\
Cedars-Sinai Medical Center \\
700 N. San Vicente Blvd., West Hollywood, CA 90069 \\
$^{\spadesuit}$\texttt{Huixin.Zhan@cshs.org} \quad
$^{\diamondsuit}$\texttt{jason.moore@csmc.edu}
}

\begin{document}

\maketitle

\begin{abstract}
Surgeons exhibit distinct operating styles shaped by training, experience, and motor behavior—yet most surgical AI systems overlook this personalization signal. We propose a novel agentic modeling approach for surgeon-specific behavior prediction in robotic surgery, combining a discrete diffusion framework with a vision-language-action (VLA) pipeline. Gesture prediction is framed as a structured sequence denoising task, conditioned on multimodal inputs including surgical video, intent language, and personalized embeddings of surgeon identity and skill. These embeddings are encoded through natural language prompts using third-party language models, allowing the model to retain individual behavioral style without exposing explicit identity. We evaluate our method on the JIGSAWS dataset and demonstrate that it accurately reconstructs gesture sequences while learning meaningful motion fingerprints unique to each surgeon. To quantify the privacy implications of personalization, we perform membership inference attacks and find that more expressive embeddings improve task performance but simultaneously increase susceptibility to identity leakage. These findings demonstrate that while personalized embeddings improve performance, they also increase vulnerability to identity leakage, revealing the importance of balancing personalization with privacy risk in surgical modeling. Code is available at: \url{https://github.com/huixin-zhan-ai/Surgeon_style_fingerprinting}. 
\end{abstract}

\begin{figure*}[!t]
    \centering
    \includegraphics[width=\linewidth]{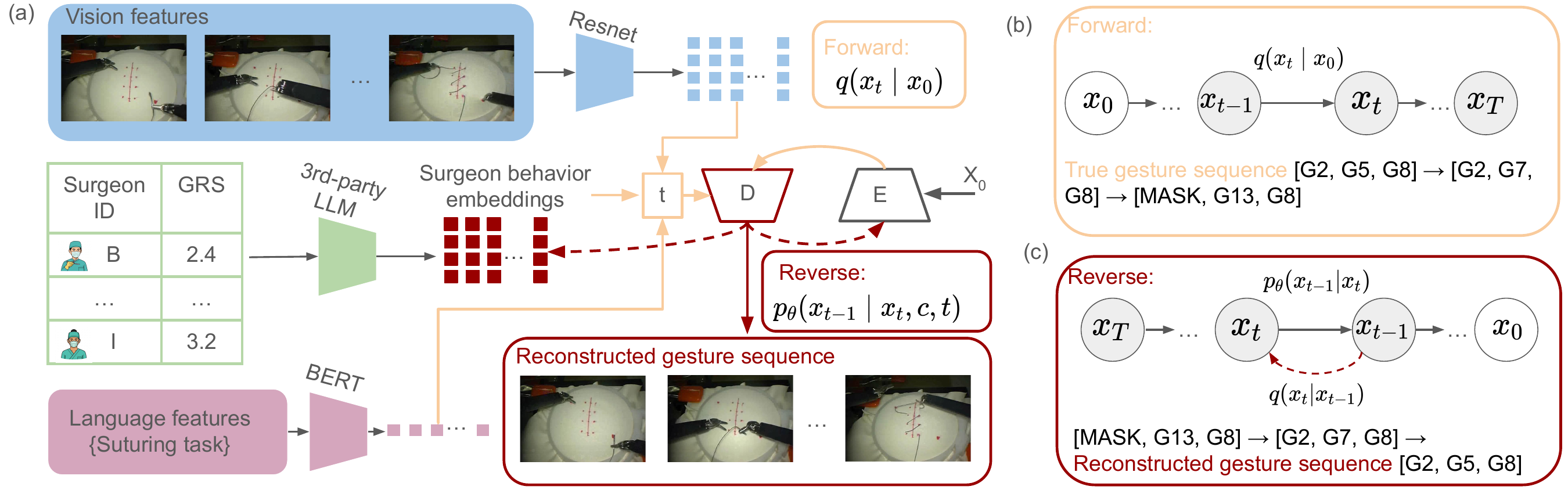}
    \caption{
       \textbf{Personalized gesture sequence prediction using diffusion models.}
(a) Vision features are extracted from surgical videos using ResNet, while task-level semantic cues are obtained from the language prompt (e.g., `suturing task'') using BERT. Surgeon-specific behavior embeddings are generated by encoding the surgeon ID and GRS using a third-party large language model (LLM). These visual, linguistic, and behavioral signals are fused to condition the denoising model. 
(b) In the forward process, categorical noise is gradually applied to the ground truth gesture sequence $x_0$ via $q(x_t \mid x_0)$, transforming discrete gestures (e.g., [G2, G5, G8]) into corrupted sequences (e.g., [MASK, G13, G8]). 
(c) The reverse process learns to reconstruct the original sequence from noise by predicting $p_\theta(x_{t-1} \mid x_t, c, t)$, where $c$ denotes multimodal context. This framework enables personalized and condition-aware gesture generation.}
    \label{fig:diffusion_fingerprint}
\end{figure*}

\begin{table*}[ht]
\centering
\caption{
Performance comparison of gesture prediction models under different surgeon identity representations. 
The \textbf{third-party LLM (ID + GRS)} setting encodes both the surgeon ID and averaged objective GRS using an LLM.
The \textbf{third-party LLM (ID only)} setting encodes only the surgeon ID using the same language model, while the \textbf{non-private baseline} uses standard learnable embeddings.
Despite similar predictive performance, the personalized LLM-based embeddings introduce higher re-identifiability risk, demonstrating a trade-off between personalization and privacy.
}

\label{tab:surgeon_fingerprint_comparison}
\begin{tabular}{lcccc}
\toprule
\textbf{Task} &\textbf{Surgeon Representation} & \textbf{Top-1 Accuracy} & \textbf{Top-5 Accuracy} & \textbf{Weighted F1-Score} \\
\midrule
\multirow{3}{*}{Suturing} &Third-party LLM (ID + GRS)            & \textbf{0.8389}$\uparrow$ & 0.9984$\downarrow$ & \textbf{0.8447}$\uparrow$ \\
&Third-party LLM (ID only)           & 0.8327 & \textbf{0.9988} & 0.8324 \\
&Non-private baseline & 0.8240 & 0.9968 & 0.8237 \\
\bottomrule
\end{tabular}
\end{table*}


\section{Introduction}
\label{sec:intro}

Personalized modeling of surgical behavior has the potential to improve intraoperative decision support, skill assessment, and robot-assisted training. In particular, fine-grained prediction of surgical gestures—low-level action primitives such as “grasp needle” or “position needle”—can reveal patterns in technique that vary across surgeons, tasks, and experience levels. Recent work~\cite{van2021gesture} has leveraged vision and language models for surgical understanding, but most existing systems remain surgeon-agnostic, failing to capture individual variation critical for personalization and performance benchmarking.

Modeling this variation, however, raises unique challenges: How can models account for stylistic behavioral differences between surgeons? Can we learn surgeon-specific representations without exposing identity information? And how can we build robust predictive systems that generalize across users while still capturing meaningful intra-surgeon variability?


To address these challenges, we introduce a discrete diffusion-based framework for personalized gesture sequence modeling in robotic surgery, grounded in a vision-language-action (VLA) paradigm. We formulate gesture prediction as a structured denoising process over discrete tokens, where the reverse diffusion model is \textit{agentically} conditioned on multimodal inputs: (i) visual context from surgical video, (ii) task-level language cues representing surgical intent, and (iii) surgeon-specific behavioral fingerprints, encoded via natural language prompts. These prompts integrate both the surgeon’s ID and their averaged Global Rating Score (GRS~\citep{gray1996global}), and are embedded using frozen third-party large language models (LLMs). The resulting embeddings are trained end-to-end with the VLA-conditioned diffusion model, enabling adaptive and personalized gesture generation. Although identity and skill cues are provided during prompt construction, they are abstracted through frozen third-party LLMs and not directly exposed to the gesture prediction module, thereby offering a degree of privacy while supporting rich personalization. 

We evaluate our method on the JIGSAWS dataset~\citep{gao2014jhu} and show that surgeon-specific embeddings improve structured sequence denoising performance. To assess potential privacy risks introduced by personalization, we conduct membership inference attacks targeting the learned embeddings. Our results reveal that while LLM+GRS embeddings enhance personalization, they also increase susceptibility to re-identification, highlighting a trade-off between behavioral fidelity and privacy. This work underscores the need for careful evaluation of privacy leakage in personalized surgical AI systems.

Our key contributions are:
(1) a diffusion-based formulation for personalized discrete gesture prediction in robotic surgery,  
(2) a language-model-based embedding scheme that encodes surgeon identity and skill cues via natural language prompts via third-party large language models,  
(3) a quantitative privacy analysis using membership inference attacks to evaluate identity leakage, and  
(4) comprehensive experiments regarding both gesture prediction accuracy and stylistic coherence on the JIGSAWS dataset demonstrating both improved personalization and the trade-offs between performance and privacy. Together, these components define an agentic modeling framework that captures context- and identity-aware behavior representation, enabling personalized gesture sequence generation and advancing the development of adaptive, privacy-conscious AI systems for surgical robotics.

\section{Method}

We formulate surgeon gesture prediction as a discrete denoising diffusion process. Each gesture sequence \( x_0 \in \mathcal{G}^T \), where \( \mathcal{G} = \{1, \ldots, K\} \) is the gesture vocabulary, is corrupted over time via a multinomial noise process. The goal is to learn a conditional model that reconstructs \( x_0 \) from a noisy sequence \( x_t \sim q(x_t \mid x_0) \), given multimodal context.

\subsection{Forward Process: Discrete Multinomial Corruption}

We define a time-indexed transition matrix \( Q_t \in \mathbb{R}^{K \times K} \) that governs the noise schedule. Each diagonal entry \( Q_t(i, i) \) decays linearly as:
\[
Q_t(i, i) = 1 - \frac{t+1}{T}, \quad Q_t(i, j \ne i) = \frac{1 - Q_t(i, i)}{K - 1}
\]
This process flips gesture tokens at increasing rates as \( t \) progresses, simulating categorical corruption.

\subsection{Reverse Process: Personalized Gesture Denoising}

We train a transformer-based model~\citep{vaswani2017attention} to approximate the reverse conditional distribution \( p_\theta(x_{t-1} \mid x_t, c, t) \), where the conditioning context \( c \) includes:

- Vision features extracted from endoscopic video using a ResNet encoder~\citep{he2016deep},

- Language features derived from a BERT encoding~\citep{devlin2019bert} of the task prompt (e.g., “suturing task”),

- Surgeon embedding \( \mathbf{s}_i \in \mathbb{R}^d \) that encodes stylistic behavioral information,

- Timestep embedding \( t \). These components are fused and injected into the denoising model at each timestep.

\subsection{Surgeon Embeddings and Personalization}


To model inter-surgeon variation, each surgeon is assigned an embedding vector \( \mathbf{s}_i \). These embeddings are either learned directly through a trainable embedding layer or derived from natural language prompts—such as ``Surgeon ID: i, GRS: 3.4"—using a frozen third-party language model (e.g., Sentence-BERT~\citep{reimers2019sentence} or MiniLM~\citep{wang2020minilm}). By conditioning on these embeddings, the model learns to denoise gesture sequences in a surgeon-specific manner, enabling stylistic profiling and personalized gesture generation. After training, the resulting embeddings can be used for downstream applications such as clustering, retrieval, or surgeon re-identification.

\begin{figure*}[ht]
    \centering
    \includegraphics[width=0.48\textwidth]{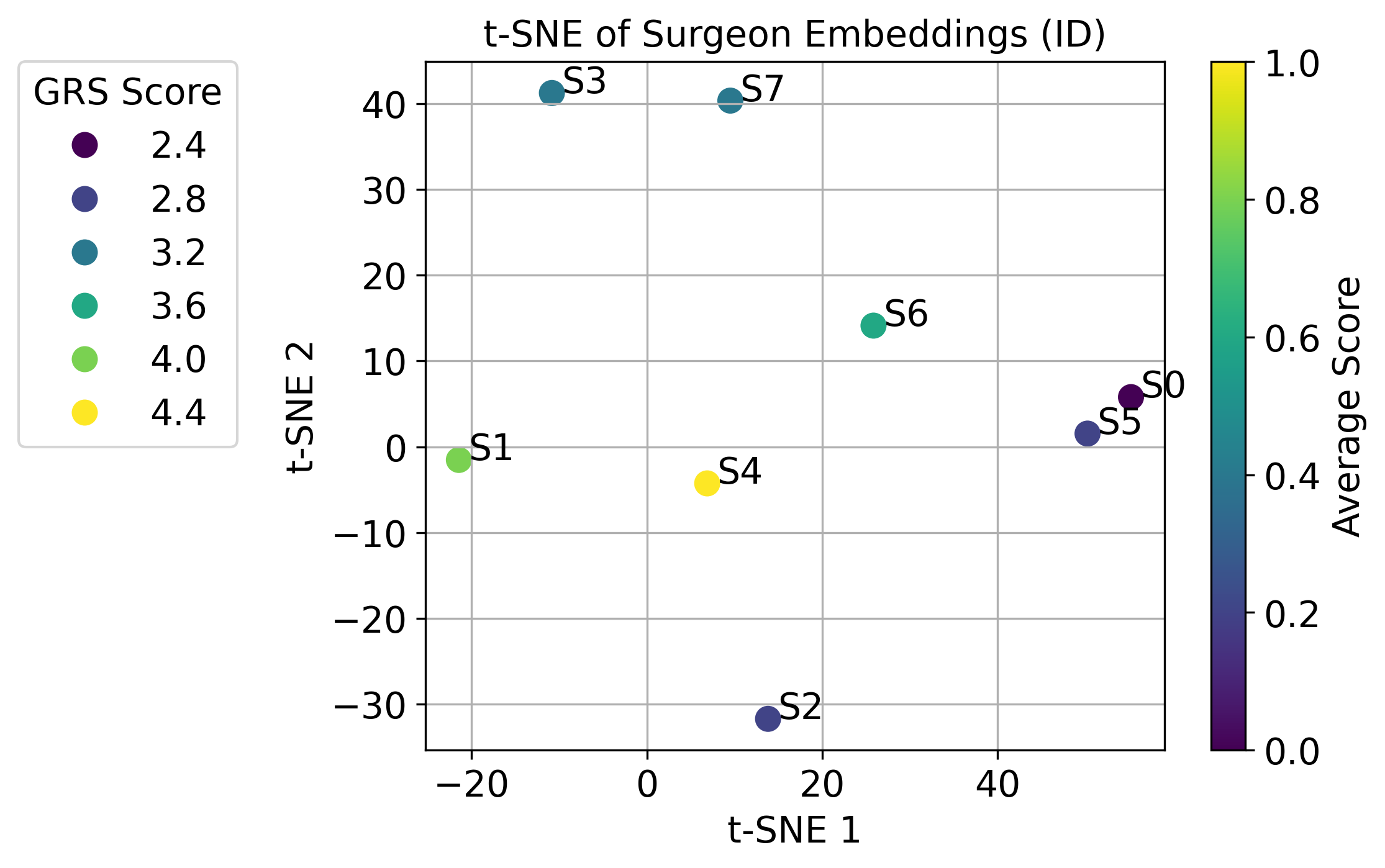}
    \includegraphics[width=0.5\textwidth]{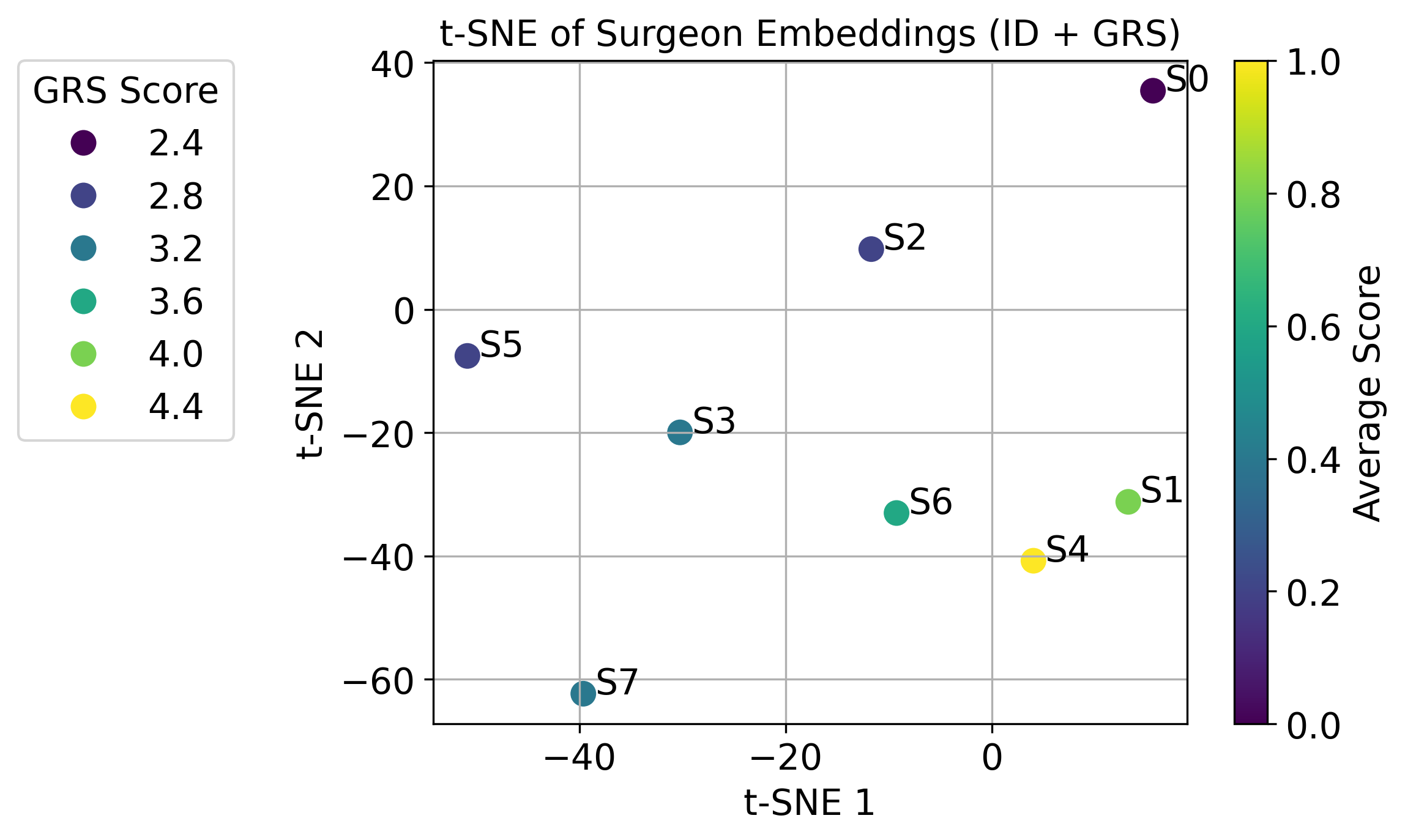}
    \caption{t-SNE of surgeon embeddings. (\textbf{Left}) Trained using surgeon ID only. (\textbf{Right}) Trained with both ID and GRS supervision. Colors indicate average skill score (GRS).}
    \label{fig:tsne}
\end{figure*}

\begin{table*}[!htbp]
\caption{
\textbf{Privacy evaluation via membership inference attack.} 
We evaluate privacy leakage under three surgeon embedding strategies by training a binary classifier to determine whether an embedding came from the training set.
Higher AUC and precision indicate greater vulnerability to identity leakage. The \textbf{third-party LLM (ID + GRS)} strategy enables stronger personalization, but at the cost of higher re-identifiability risk.
}
\label{tab:membership_comparison}
\centering
\begin{tabular}{lcccccc}
\toprule
\textbf{Task} &\textbf{Surgeon Representation} & \textbf{Accuracy} & \textbf{Precision} & \textbf{Recall}  & \textbf{F1 Score} & \textbf{AUC} \\
\midrule
\multirow{3}{*}{Suturing} 
&Third-party LLM (ID + GRS)            & \textbf{0.997}$\uparrow$ & \textbf{0.998}$\uparrow$ & \textbf{1.000}$\uparrow$ & \textbf{0.988}$\uparrow$ & \textbf{1.000}$\uparrow$ \\
&Third-party LLM (ID only)             & 0.889 & 0.500 & 1.000 & 0.667 & 1.000 \\
&Non-private baseline   & 0.889 & 0.000 & 0.000 & 0.000 & 0.469 \\
\bottomrule
\end{tabular}
\end{table*}


\subsection{Training Objective}

We sample a timestep \( t \sim \mathcal{U}(0, T) \) uniformly and compute a noisy gesture sequence \( x_t \sim q(x_t \mid x_0) \). The model is trained to predict the original gesture \( x_0 \) using a cross-entropy loss:
\[
\mathcal{L}_{\text{CE}} = \mathbb{E}_{x_0, t, x_t} \left[ - \log p_\theta(x_0 \mid x_t, c, t) \right]
\]
The loss is computed over all gesture positions, and gradients are backpropagated through the entire diffusion trajectory.

\subsection{Inference}

At test time, we formulate gesture prediction as a generative inference task over discrete sequences. Starting from a fully corrupted gesture sequence \( x_T \sim q(x_T \mid x_0) \), which represents near-uniform categorical noise, the model applies the learned reverse process iteratively from \( t = T \) to \( 0 \). This produces a denoised sequence \( \hat{x}_0 \) that reconstructs the likely gesture trajectory under the current visual, linguistic, and behavioral context.

This formulation departs from traditional gesture recognition methods, which typically rely on direct sequence classification or frame-wise decoding. Instead, we cast inference as a structured sequence generation problem conditioned on multimodal and surgeon-specific embeddings. The process is inherently probabilistic and allows for uncertainty modeling and sample diversity, enabling the system to generate plausible personalized behaviors.

We evaluate inference performance along two key axes: 
(i) \textit{gesture prediction accuracy}, which quantifies the model's ability to reconstruct the ground truth gesture sequence, and 
(ii) \textit{stylistic coherence}, which assesses how closely the generated trajectory reflects the unique behavioral fingerprint of the target surgeon, as captured by their personalized embedding.




\subsection{Privacy Risk Analysis via Membership Inference}

To quantify the privacy risk associated with personalized embeddings, we conduct a membership inference attack~\citep{shokri2017membership}. Given the learned surgeon embeddings \( \{\mathbf{s}_i\} \), we simulate an adversary that trains a binary classifier (e.g., XGBoost~\citep{chen2016xgboost}) to distinguish between in-training (member) embeddings and synthetic out-of-distribution (non-member) ones. High classifier performance—measured by AUC, accuracy, and F1-score—indicates a greater risk of identity leakage.

We compare privacy vulnerability across three embedding strategies. As reported in Section~\ref{sec:experiments}, the LLM (ID + GRS) embedding achieves the best gesture prediction performance, but also exhibits the highest susceptibility to membership inference attacks (AUC = 1.000). This reveals a key privacy-performance trade-off: richer personalization yields more discriminative embeddings, which are easier to link back to training data. In contrast, the non-private baseline produces less structured embeddings that offer weaker personalization but lower risk under attack.

These findings emphasize the need for explicit privacy evaluation when deploying personalized models in clinical settings, especially when embedding behavioral signals like skill.

\section{Experiments}
\label{sec:experiments}
\noindent

To evaluate the effectiveness of our proposed framework, we design a series of experiments aimed at answering the following key scientific questions:

\begin{itemize}
    \item \textbf{Q1:} How do different surgeon representation strategies (e.g., learnable ID embeddings, third-party LLMs (ID only), and third-party LLMs (ID + GRS)) impact personalized gesture sequence modeling?
    
    \item \textbf{Q2:} Can natural language embeddings that incorporate clinically validated metrics (e.g., GRS) enable personalization?
    
    \item \textbf{Q3:} Do the learned surgeon embeddings capture meaningful and structured variation in skill or behavior across different individuals?
    
    \item \textbf{Q4:} To what extent do different personalization methods expose the model to privacy leakage under adversarial conditions such as membership inference attacks?
\end{itemize}

\paragraph{Settings} We design ablation experiments on surgeon embedding strategies, heatmap visualizations, GRS-informed fingerprinting analyses, and privacy evaluations to systematically investigate each of these research questions. The model is trained for 20 epochs using the Adam optimizer with a learning rate of $1 \times 10^{-3}$ and a batch size of 32. All projection layers, surgeon and timestep embeddings, and the denoising network operate in a shared hidden space of size 512. Visual representations are extracted from ResNet and projected from a 1000-dimensional output space. Textual prompts (e.g., ``suturing task”) are encoded using BERT, yielding 768-dimensional embeddings corresponding to the [CLS] token. The discrete gesture space comprises 16 tokens, including 15 surgical gestures (G1–G15) and a dedicated [MASK] token for noise modeling. Each input sequence consists of 5 discrete gesture tokens, capturing short temporal contexts within a surgical trial. To enable structured denoising, we apply a forward corruption process over 10 discrete timesteps, during which categorical noise is progressively introduced to the sequence. 

The results and insights are presented in the following sections.
\subsection{Ablation on Surgeon Embedding Strategies.}
To assess the impact of surgeon-specific conditioning on gesture prediction, we compare three distinct embedding strategies:

\begin{enumerate}
    \item \textbf{Non-private baseline:} We use a learnable embedding layer that maps each surgeon ID to a unique vector. This approach directly exposes identity and serves as an upper-bound reference in terms of capacity and specificity.

    \item \textbf{Third-party LLM (ID only):} Surgeon IDs are formatted into short natural language prompts (e.g., ``Surgeon ID: 3'') and encoded using a publicly available sentence-level language model (e.g., MiniLM). This embedding is kept fixed and projected to the model's hidden dimension via a learnable linear layer.

\item \textbf{Third-party LLM (ID + GRS):} To incorporate behavioral priors, we construct natural language prompts that include both the surgeon ID and their averaged GRS, e.g., ``Surgeon ID: 3, average skill score: 3.75''. These prompts are encoded using a frozen third-party language model (e.g., Sentence-BERT), and projected into the model’s hidden space. This strategy captures personalized style in a clinically grounded manner, while avoiding direct integration of raw identifiers into the gesture prediction model.

\end{enumerate}

We evaluate all strategies within the same diffusion-based gesture prediction framework, under identical training and testing conditions. As shown in Table~\ref{tab:surgeon_fingerprint_comparison}, the third-party LLM (ID + GRS) method achieves the highest Top-1 accuracy (83.89\%) and weighted F1-score (0.8447), with near-perfect Top-5 accuracy (99.84\%), outperforming the ID-only and non-private baselines. These results reflect the benefit of incorporating clinically grounded behavioral cues like GRS into surgeon representations.

However, our privacy analysis reveals that this gain in personalization comes at the cost of increased identity leakage under membership inference attacks. This finding reveals a core tension in surgical modeling: richer behavioral embeddings improve prediction, but also make models more susceptible to privacy breaches. Therefore, while third-party LLM embeddings with GRS offer strong personalization capabilities, they also demand careful consideration of privacy risk.

\subsection{Qualitative Analysis via Gesture Distribution Heatmaps}



To further understand the behavioral distinctions captured by each surgeon embedding strategy, we visualize the predicted gesture distributions across surgeons using heatmaps. Each heatmap cell represents the frequency of a predicted gesture token (e.g., G1--G15) for a specific surgeon, aggregated across all test sequences.

Figure~\ref{fig:gesture_heatmaps} compares the predicted gesture distributions under three settings: (a) the non-private baseline using learnable ID embeddings, (b) the third-party LLM (ID only) condition, and (c) the third-party LLM (ID + GRS) embedding. These visualizations qualitatively reinforce our quantitative results: the third-party LLM (ID + GRS) variant achieves personalization effects comparable to the non-private baseline—despite not relying on direct access to learned identity embeddings.

\subsection{Surgeon Fingerprinting and GRS Association}

To explore whether our model captures meaningful variations in surgical behavior, we visualize surgeon embeddings using t-SNE. Each point in Figure~\ref{fig:tsne} represents an individual surgeon, and colors indicate their average GRS performance score.

The left panel shows embeddings learned when the model is only provided surgeon IDs. In this setting, the embedding space primarily reflects identity but does not exhibit clear separation by skill level. In contrast, the right panel shows embeddings learned with both surgeon ID and GRS supervision. Here, surgeons with similar skill levels are more coherently grouped, and the embeddings demonstrate smoother gradients along the GRS axis.

This suggests that our personalized representation — guided by visual, linguistic, and behavioral cues — can encode latent skill information in an unsupervised or weakly-supervised fashion. Embedding models that integrate clinical supervision like GRS may enhance both interpretability and downstream personalization.

\subsection{Privacy Evaluation via Membership Inference}

We assess privacy leakage from different surgeon embedding strategies using a membership inference attack, in which an adversary attempts to determine whether a given embedding originated from the model’s training data. Results are shown in Table~\ref{tab:membership_comparison}.

The \textbf{third-party LLM (ID + GRS)} representation is the most expressive, but also the most vulnerable. It enables highly personalized gesture modeling, but the resulting embeddings are easily distinguishable from synthetic (non-member) embeddings, with an AUC of 1.000, precision of 0.998, and F1-score of 0.988. This indicates that the embedding space reveals strong identity-linked signals, raising concerns about privacy leakage.

The \textbf{third-party LLM (ID only)} strategy also exhibits high recall and AUC, but its lower precision (0.500) and F1-score (0.667) suggest a less precise separation between members and non-members. This reflects a weaker entanglement of identity information compared to the GRS-conditioned variant.

The \textbf{non-private baseline} shows low attack performance, with zero precision and recall and low AUC (0.469). While this may seem desirable from a privacy standpoint, it more likely reflects noisy or unstructured embeddings rather than true robustness.

Overall, these results highlight a trade-off: more expressive, personalized embeddings improve task performance but increase susceptibility to identity inference attacks. This underscores the importance of quantifying privacy risks when designing surgeon-specific models in surgical AI.

\section{Discussion}
Robotic surgical systems are increasingly being integrated with multimodal learning pipelines to enhance automation, guidance, and assessment. However, most existing approaches overlook individual behavioral variability and the implications of modeling personalization. In this section, we organize our discussion around three key pillars that shape the landscape of personalized surgical modeling: (1) current VLA systems, (2) surgeon fingerprinting for behavior encoding, and (3) privacy implications of personalization.
\subsection{Current Vision-Language-Action (VLA) Systems}

Recent advances in surgical AI have explored multimodal learning through vision-language-action (VLA) frameworks \citep{zargarzadeh2025decision,ding2025visual}. These models integrate visual and textual signals to understand surgical phases and gestures. However, most existing systems adopt a one-size-fits-all architecture that models behavior agnostic to individual variation. Works like \citep{dipietro2019segmental,ma2024transsg,kiyasseh2023vision} focus on phase prediction or gesture segmentation, but assume shared behavior across surgeons. As a result, they may generalize poorly in contexts requiring personalization, such as training simulators or adaptive feedback systems.

\subsection{Surgeon Style Fingerprinting}

Few prior studies explicitly model stylistic differences across surgeons. Earlier efforts such as \citep{zia2018automated, funke2019video} attempt to correlate motion patterns with skill, but do not offer a generative formulation for personalized prediction. Our work builds on the idea of behavioral fingerprinting by embedding per-surgeon style representations and generating gesture sequences via discrete diffusion. This complements prior gesture forecasting approaches \citep{shi2022recognition,honarmand2024vidlpro} with a focus on personalized generation. The ability to encode style in a structured embedding opens new possibilities for re-identification, clustering, and personalized curriculum learning.

\subsection{Privacy Quantification via Membership Inference}

While privacy has received growing attention in medical ML \citep{dwork2014algorithmic, shokri2017membership}, few works examine privacy risks in behavioral embeddings. Most privacy studies focus on image reconstruction or record linkage \citep{fredrikson2015model}, not personalized sequence generation. Our membership inference analysis reveals a tension between expressiveness and privacy: richer embeddings (e.g., LLM+GRS) perform better but are more susceptible to identity leakage. This is aligned with recent work on privacy-utility trade-offs in embedding models \citep{carlini2021extracting}. Future extensions could explore differential privacy or adversarial training to mitigate this risk.

Collectively, our findings underscore the need for models that are not only multimodal but also personalized, privacy-aware, and behaviorally grounded. Our framework advances the direction of agentic AI by enabling systems to adapt to user identity, intent, and context. By capturing surgeon-specific behavioral fingerprints and reasoning over structured gesture sequences, we take a step toward building surgical agents capable of individualized learning, proactive assistance, and trust-aware interaction in clinical environments.

{
    \small
    \bibliographystyle{ieeenat_fullname}
    \bibliography{main}
}

\setcounter{page}{1}
\onecolumn
\begin{figure}[!htbp]
    \centering
    \begin{subfigure}{0.9\linewidth}
        \centering
        \includegraphics[width=\linewidth]{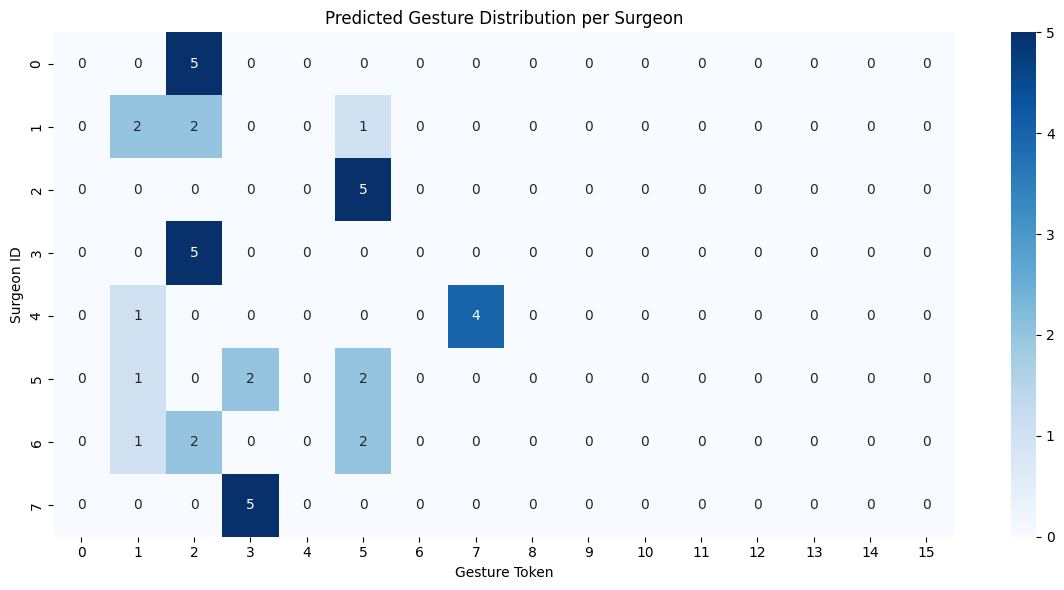}
        \caption{Non-private baseline}
    \end{subfigure}
        \hfill
        \begin{subfigure}{0.9\linewidth}
        \centering
        \includegraphics[width=\linewidth]{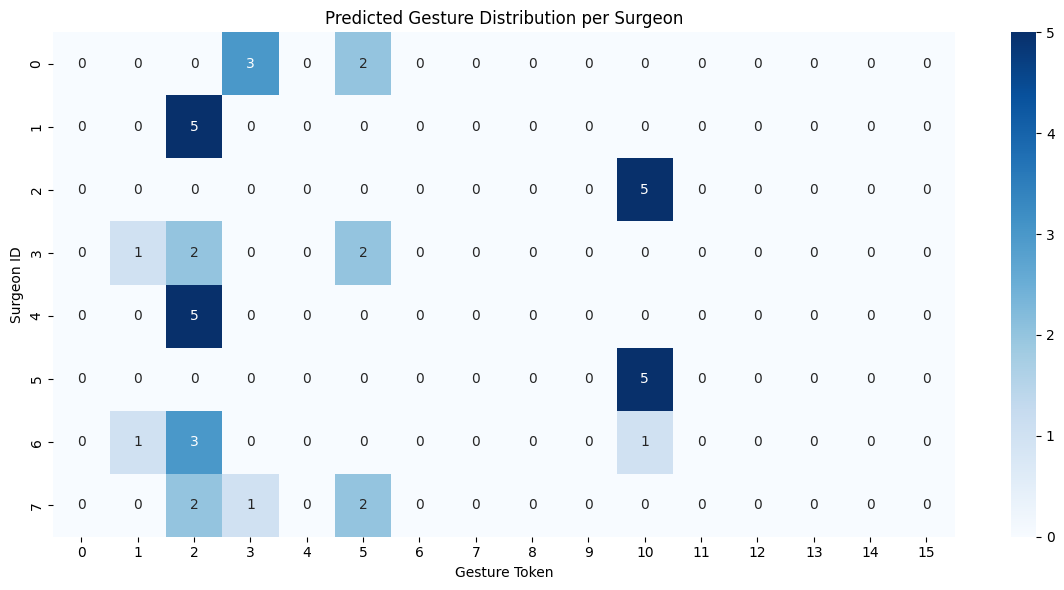}
        \caption{Third-party LLM (ID only)}
    \end{subfigure}
        \caption{
        Predicted gesture token distributions across surgeons.
        Rows represent surgeon IDs and columns denote gesture tokens.
    }
    \label{fig:gesture_heatmaps}
\end{figure}


\begin{figure*}[!htbp]
    \ContinuedFloat
    \centering
    \begin{subfigure}{0.9\linewidth}
        \centering
        \includegraphics[width=\linewidth]{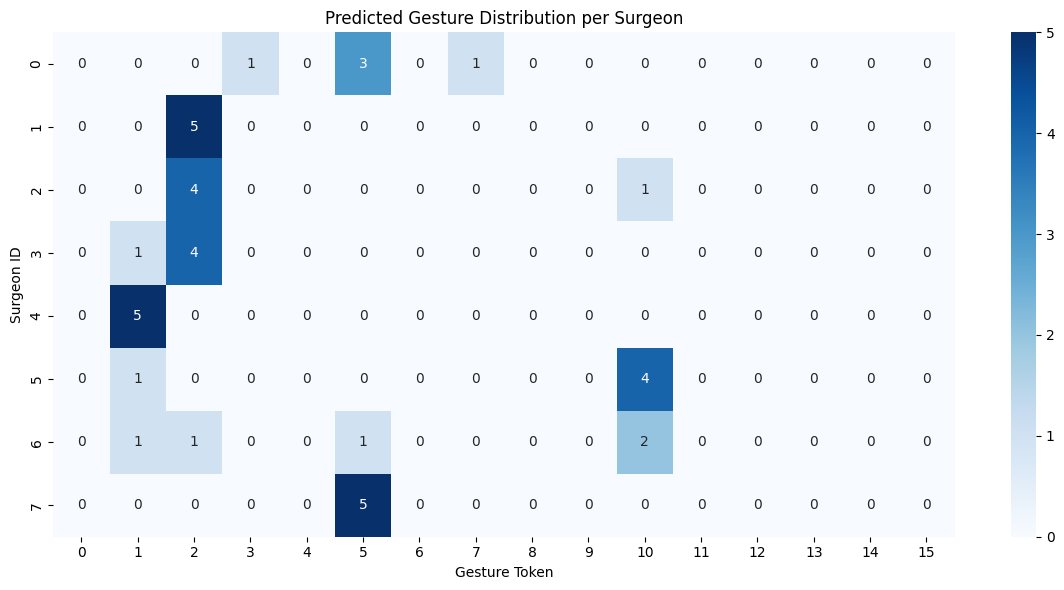}
        \caption{Third-party LLM (ID + GRS)}
    \end{subfigure}
    \caption{(continued) Panel (c) shows structured, personalized distributions under ID + GRS embedding.}
\end{figure*}

\end{document}